\documentclass[final,3p,times,twocolumn]{elsarticle}

\usepackage{amssymb}
\usepackage{amsmath}
\usepackage{booktabs}
\usepackage{color}
\usepackage{multirow}
\usepackage{array}
\usepackage{rotating}
\usepackage[table]{xcolor}
\usepackage{algorithm,algorithmic}

\journal{Pattern Recognition}

\begin{document}

\begin{frontmatter}



\title{DyGLNet: Hybrid Global-Local Feature Fusion with Dynamic Upsampling for Medical Image Segmentation}


\author[1]{Yican Zhao}
\author[2]{Ce Wang}
\author[3]{You Hao}
\author[1]{Lei Li}
\author[1]{Tianli Liao\corref{cor1}}

\cortext[cor1]{Corresponding author. E-mail address: tianli.liao@haut.edu.cn.}

 \address[1]{College of Information Science and Engineering, Henan University of Technology, Zhengzhou, 450001, China}
 \address[2]{School of Science, Sun Yat-sen University, Shenzhen, Guangdong, 518107, China}
 \address[3]{Key Lab of Intelligent Information Processing of Chinese Academy of Sciences (CAS), Institute of Computing Technology, CAS, Beijing 100190, China}


\begin{abstract}
Medical image segmentation grapples with challenges including multi-scale lesion variability, ill-defined tissue boundaries, and computationally intensive processing demands. This paper proposes the DyGLNet, which achieves efficient and accurate segmentation by fusing global and local features with a dynamic upsampling mechanism. The model innovatively designs a hybrid feature extraction module (SHDCBlock), combining single-head self-attention and multi-scale dilated convolutions to model local details and global context collaboratively. We further introduce a dynamic adaptive upsampling module (DyFusionUp) to realize high-fidelity reconstruction of feature maps based on learnable offsets. Then, a lightweight design is adopted to reduce computational overhead. Experiments on seven public datasets demonstrate that DyGLNet outperforms existing methods, particularly excelling in boundary accuracy and small-object segmentation. Meanwhile, it exhibits lower computation complexity, enabling an efficient and reliable solution for clinical medical image analysis. The code will be made available soon.
\end{abstract}




\end{frontmatter}



\section{Introduction}\label{sec:intro}


Medical image segmentation plays a crucial role in numerous downstream computer-aided intervention tasks, including disease diagnosis, treatment planning, and surgical navigation. Practically, the process involves the precise delineation of anatomical structures or pathological regions within medical images, making it a cornerstone for advancing personalized medicine and patient outcomes. Recently, with the increasing necessity for modeling hundreds of tissues~\cite{wasserthal2023totalsegmentator}, efficient segmentation models have become demanded~\cite{hatamizadeh2021swin,yu2023unest}.

In the past few years, witnessing the significant advantages of deep learning in vision tasks, convolutional neural networks (CNNs) have been introduced and have dominated medical image segmentation due to their superior performance. Among them, the UNet~\cite{ronneberger2015u} and its variants have attracted researchers and been widely employed. Typically, the ``U-shape" models consist of encoder-decoder structures for representation learning and pixel-wise segmentation, respectively. However, such CNN-based models can hardly capture and model global representation because of their localized receptive fields~\cite{vaswani2017attention}. To extract global representation and model the long-range dependencies, the Vision Transformers (ViT)~\cite{dosovitskiy2020image} have been introduced and demonstrated exceptional representational learning capability and effectiveness in vision and medical image applications~\cite{zhou2021nnformer,hatamizadeh2022unetr}. In general, ViTs tokenize images into 1D sequences and leverage self-attention blocks to enable communications among the whole image instead of utilizing large convolutional kernels, benefiting dense prediction tasks. Nevertheless, the core self-attention mechanism requires highly complex computation, limiting its efficiency in clinical deployment.

To leverage both advantages, researchers~\cite{zhang2021transfuse,meng2021exploiting,valanarasu2021medical,IQBAL2025111028,HUANG2025111274,YANG2026112139} thus propose hybrid structures to combine CNNs and ViTs as distinct encoder branches that both extract representational features, followed by fusion and connecting to the decoder. Such a design takes into account both the global and local information in representation learning and consecutively stacks them. Moreover, several self-supervised works are further proposed to enhance the robustness and efficiency, rendering representation learning without manual labels.

Although these technologies enable models to achieve better performances in medical image segmentation to a certain extent, they neglect in-depth analysis between the model design and data characteristics, limiting the potential improvement on specific tasks. We next thoroughly analyze the problem and provide subsequent modifications from the perspective of the analysis. Specifically, medical image segmentation (especially the polyp, dermoscopic, and pathological images) still confronts three critical challenges in practical applications: 1) lesions appear \textbf{significant differences in shape and size}, which renders traditional methods with single receptive fields (e.g., CNNs with fixed kernel size and ViTs with fixed patch size) ineffective for multi-scale feature extraction; 2) there exist \textbf{blurred boundaries between lesions and normal tissues} (e.g., low-contrast polyp edges), making it difficult to restore segmentation contours accurately. This leads to edge blurring during the decoder process with traditional upsampling strategies (e.g., bilinear), and transposed convolutions instead suffer from high computation and are prone to introducing checkerboard artifacts; 3) ViTs and hybrid models are with \textbf{high model computational complexity}, restricting algorithm deployment in resource-constrained scenarios such as mobile or edge devices.

To address the above challenges, we propose a novel medical image segmentation network \textbf{Dy}namic \textbf{G}lobal-\textbf{L}ocal \textbf{Net}work, named DyGLNet. Facing various shapes and scales, we firstly propose to collaboratively model both global and local features with SHDCBlock in the encoding process. Then we utilize the DyFusionUp module in the decoding process to guarantee semantic consistency and structural accuracy, which helps deal with blurred boundaries. Further targeting to enhance model efficiency, the lightweight dynamic adaptive upsampling module is later integrated, highly reducing model complexity.
Based on such a lightweight design, DyGLNet integrates local detail perception and global context modeling capabilities, effectively enhancing segmentation accuracy and deployment efficiency, as shown in Fig.~\ref{fig:1}. The main contributions are as follows:

\begin{figure}
	\centering
	\includegraphics[width=\linewidth]{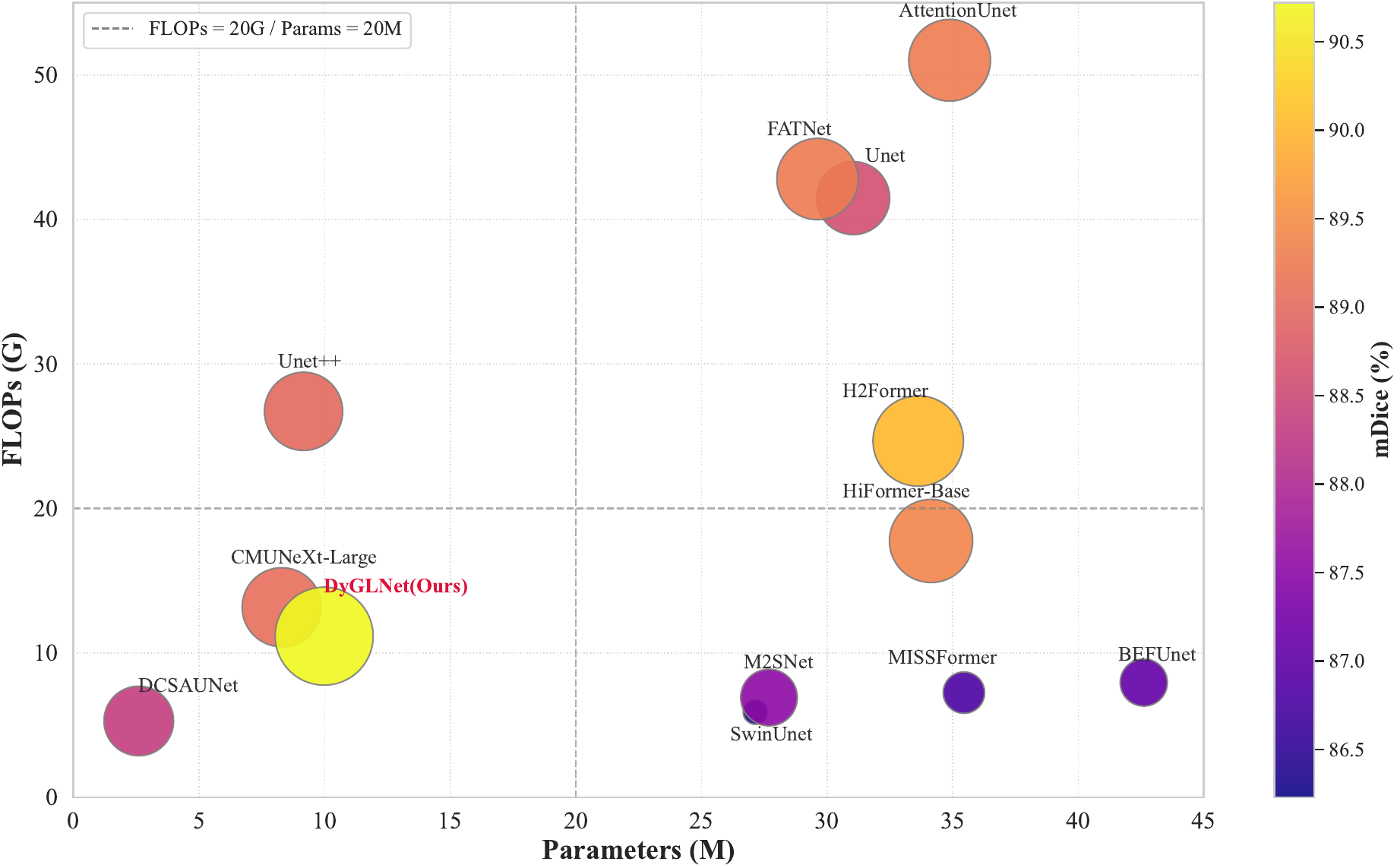}
	\caption{Visualization of model trade-off between computational cost (FLOPs and parameters) and segmentation accuracy. Bubble size and color represent the average Dice score (mDice) across seven public datasets. Our proposed DyGLNet achieves the best accuracy with significantly lower computational cost.}
	\label{fig:1}
\end{figure}

\begin{enumerate}
	\item To enhance the capability of representing multi-scale lesions and modeling long-range dependencies, we propose a hybrid feature extraction structure, named SHDCBlock, which consists of a single-head attention mechanism and multi-scale dilated convolutions.
	\item The lightweight dynamic adaptive upsampling module, DyFusionUp, is introduced. With learnable spatial offset parameters for intelligent feature map reconstruction, our module significantly improves boundary restoration accuracy and small-object recognition accuracy.
	\item Comprehensive experiments are conducted on seven public medical image segmentation datasets. The results verify that DyGLNet consistently outperforms existing mainstream methods while maintaining low computational costs, achieving a better balance between segmentation accuracy and model complexity.
\end{enumerate}

\section{Related Work}\label{sec:related}

In recent years, with the advancement of deep learning, numerous methods based on CNNs and transformer architectures have been widely applied to medical image segmentation tasks, yielding remarkable performances.

\subsection{CNN-Based Methods}


The first U-shaped architecture, UNet~\cite{ronneberger2015u} was introduced, which employs a symmetric encoder-decoder structure with skip connections to fuse low-level detail and high-level semantic information. This architecture has demonstrated significant success in biomedical image segmentation tasks. Various variants based on the U-shape structure were subsequently proposed to further enhance model performance.

AttentionUNet~\cite{oktay2018attention} incorporates an attention mechanism to enhance the perception of crucial regions, while UNet++~\cite{zhou2019unet++} redesigns skip connections to achieve dense aggregation of multi-scale features. 
The UNet 3+~\cite{huang2020unet} combines full-scale skip connections and deep supervision to benefit organs of varying sizes. CPFNet~\cite{feng2020cpfnet} combines multiple global pyramid guidance modules and a scale-aware pyramid fusion module to gradually exploit and fuse rich context information. 
nnUNet~\cite{isensee2021nnu} develops an automatically configurable framework to offer cross-dataset generalization ability. 
DCSAUNet ~\cite{xu2023dcsau} presents a split attention module and constructs a deeper, more compact network structure to strengthen the fusion between low-level details and high-level semantic features. 
M2SNet~\cite{zhao2023m} proposes a general multi-scale in multi-scale subtraction network consisting of multiple multi-scale subtraction units to simultaneously capture detailed and structural cues. 
BFNet~\cite{zhan2024bfnet} introduces full-scale connections and a global context attention mechanism to better learn the positional information of segmentation targets, thereby improving the learning of boundary information. 
I2UNet~\cite{dai2024i2u} proposes a dual-path UNet enabling deep layers to learn more comprehensive features that contain both low-level detail descriptions and high-level semantic abstractions. 
CMUNeXt~\cite{tang2024cmunext} proposes an efficient, lightweight network with a skip fusion block. It leverages a large kernel and an inverted bottleneck design to extract global contextual information. 

Despite their remarkable performance, existing CNN-based methods are inherently limited to local feature extraction due to the nature of convolutional operations. This restricts their ability to comprehensively model long-range dependencies and global semantic context in images. Consequently, these methods often produce inaccurate segmentation for medical images containing large-scale structures, blurred boundaries, or complex backgrounds.


\subsection{Transformer-based Methods}


The standard transformer architecture~\cite{vaswani2017attention} in natural language processing and its variants~\cite{dosovitskiy2020image,liu2021swin,yun2024shvit} in computer vision tasks have achieved breakthroughs due to their powerful capability for modeling global and long-range information. Inspired by this success, various efforts have been devoted to introducing transformer architecture into medical image segmentation. For example, 
TransUNet~\cite{chen2021transunet} embeds ViT into the UNet structure and introduces self-attention modules in the encoder stage to achieve global information modeling. 
MISSFormer~\cite{huang2021missformer} is designed to extract long-range dependencies and local context of multi-scale features using an enhanced transformer block and a transformer context bridge. 
SwinUNet~\cite{cao2022swin} introduces a window attention mechanism to enhance context modeling capabilities while balancing computational efficiency.

For 3D medical image segmentation, 
nnFormer~\cite{zhou2021nnformer} combines interleaved convolution and self-attention operations, proposing a 3D transformer for volumetric medical image segmentation. 
Swin UNETR~\cite{hatamizadeh2021swin} uses a hierarchical Swin transformer as the encoder, projecting the multimodal input data into a 1D sequence of embeddings. 
UNetR~\cite{hatamizadeh2022unetr} adopts a U-shaped network design, utilizing a transformer as the encoder to learn sequence representations of the input volume. 
UNesT~\cite{yu2023unest} designs a simplified transformer encoder that converges faster and enables local communication between adjacent patches by aggregating them hierarchically.

Transformer-based methods also present challenges. Firstly, the computational complexity of its self-attention mechanism increases quadratically, which limits its application to high-resolution medical images. On the other hand, their capability for modeling local structures is limited, making it difficult to accurately restore edge details and small targets.

\subsection{Hybrid Architecture-based Methods}

\begin{figure*}
	\centering
	\includegraphics[width=\linewidth]{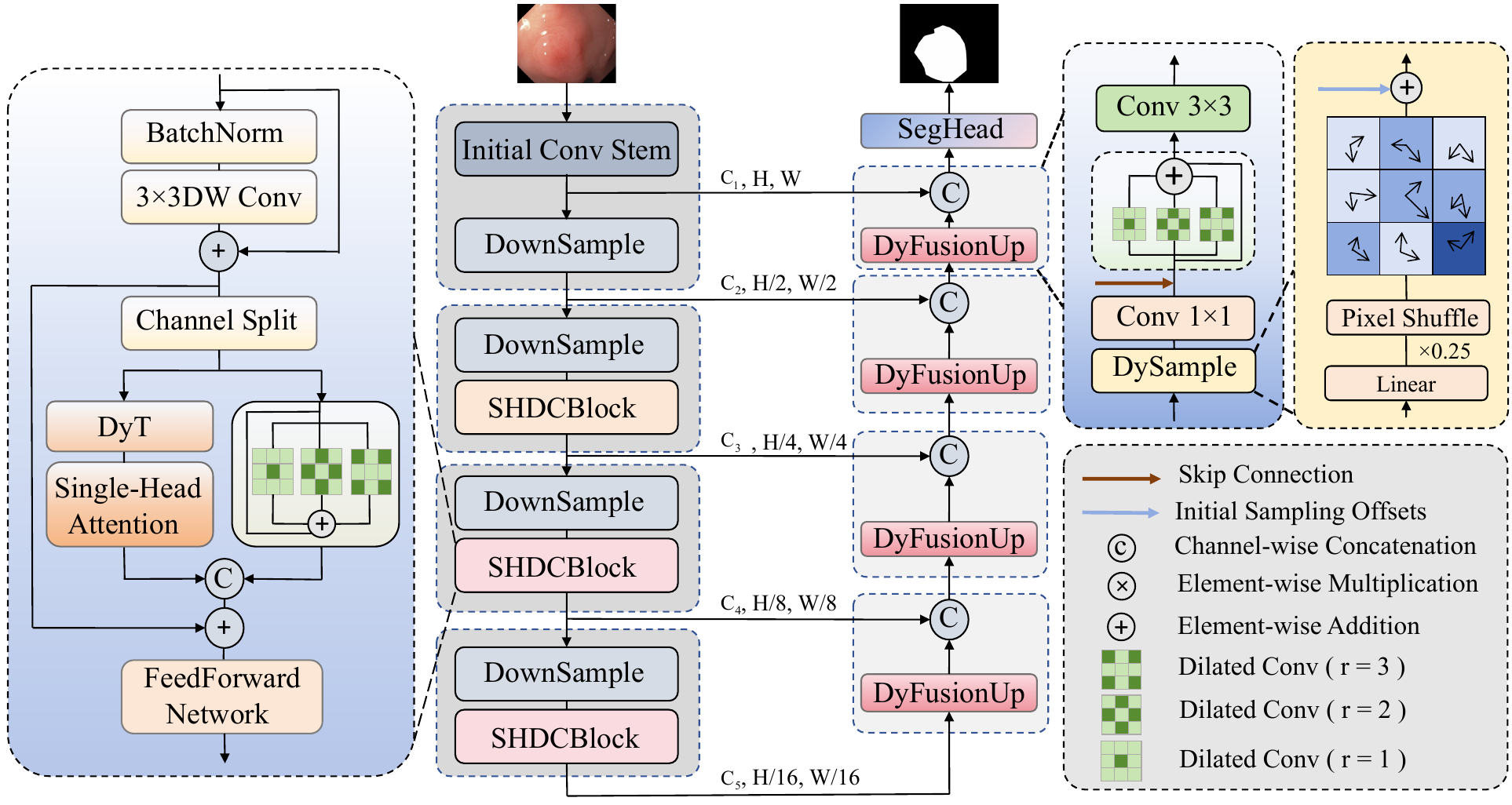}
	\caption{Overview of our DyGLNet. The model adopts a symmetric U-shaped structure, which consists of a left-side encoder, a right-side decoder, a carefully designed SHDCBlock for hybrid feature extraction, and DyFusionUp modules for adaptive upsampling. The network enables multi-scale feature interaction across various resolution levels and incorporates skip connections to facilitate spatial information recovery. The final output preserves the original spatial resolution.}
	\label{fig:pipeline}
\end{figure*}


Follow-up studies have tried to fuse CNN and transformer structures to balance local details and global semantic information. 
Transfuse~\cite{zhang2021transfuse} combines transformers and CNNs in a parallel style to capture global dependencies and low-level spatial details in a shallower manner. 
MedT~\cite{valanarasu2021medical} proposes a gated axial-attention model that introduces an extra control mechanism within the self-attention module. This model can be effectively trained using small-scale medical datasets. 
FATNet~\cite{wu2022fat} incorporates an additional transformer branch to model long-range dependencies and capture global contextual information. Meanwhile, a memory-efficient decoder and a feature adaptation module are adopted to strengthen feature fusion across adjacent-level features. 
HiFormer~\cite{heidari2023hiformer} uses the Swin Transformer module and a CNN-based encoder to design two multi-scale feature representations, while H2Former~\cite{he2023h2former} proposes a hierarchical hybrid vision transformer that can integrate local CNN information and multi-scale channel attention features of the transformer simultaneously.
BEFUnet~\cite{manzari2024befunet} proposes a novel U-shaped network comprising a local cross-attention feature module, a double-level fusion module, and a dual-branch encoder to enhance the fusion of body and edge information.
In contrast, CFFormer~\cite{li2025cfformer} introduces a cross-feature attention module to facilitate interactions between channel features, as well as an X-spatial feature fusion module to reduce differences in semantic information in spatial features.

\subsection{Upsampling Methods}

Upsampling is a critical step in restoring spatial resolution and boundary information. It directly influences the accuracy and robustness of the segmentation results. Traditional bilinear interpolation struggles to capture fine-grained boundary information in image structures. This often results in edge blurring and target degradation in the segmentation results. Transposed convolution has been incorporated into mainstream encoder-decoder architectures, such as UNet~\cite{ronneberger2015u} and CFFormer~\cite{li2025cfformer}, to enhance reconstruction quality. However, transposed convolution suffers from large parameter volumes and checkerboard artifacts, which limit its application in efficient, lightweight models.

In recent years, dynamic modeling approaches have shown remarkable superiority in upsampling tasks. 
Pixel Shuffle~\cite{shi2016real} introduces a parameter-free strategy, though its reconstruction quality heavily depends on the feature representation capability of preceding layers, thereby limiting adaptability to boundary information variations. Carafe~\cite{wang2019carafe} realizes pixel-level adaptive reconstruction via content-aware recombination mechanisms. Fade~\cite{lu2022fade} generates upsampling kernels by fusing encoder-decoder features, demonstrating strong versatility for dense prediction tasks with balanced performance and computational efficiency. 
Meanwhile, Sapa~\cite{lu2022sapa} employs local similarity measures between encoder and decoder features to synthesize upsampling kernels, excelling in detail restoration for semantic segmentation tasks. 
DySample~\cite{liu2023learning} reinterprets upsampling from point sampling theory, designing an ultra-lightweight dynamic upsampler that significantly reduces computational overhead while maintaining superior performance across multiple dense prediction scenarios.

Although existing methods have improved the upsampling quality to some extent, they still suffer from rigid structures, poor boundary reconstruction capabilities, or high computational costs. This paper, therefore, proposes a dynamic upsampling module that combines a lightweight design with a learnable offset mechanism based on DySample.

\section{Methodology}

The proposed DyGLNet, as in Fig.~\ref{fig:pipeline}, consists of a hybrid feature extraction encoder and the lightweight dynamic adaptive upsampling decoder. The main trunk is based on a lightweight Transformer-enhanced convolutional network, and multi-level residual connections are combined to achieve full-scale feature fusion. Under such considerably design, the model is capable of effectively fusing multi-scale contextual information, finely restoring lesion boundaries, and further achieving the balance between computational efficiency and segmentation performance. Next, we introduce each included module in detail.

\subsection{Overall Encoder-Decoder Structure}

\begin{figure}[t]
	\centering
	\includegraphics[width=\linewidth]{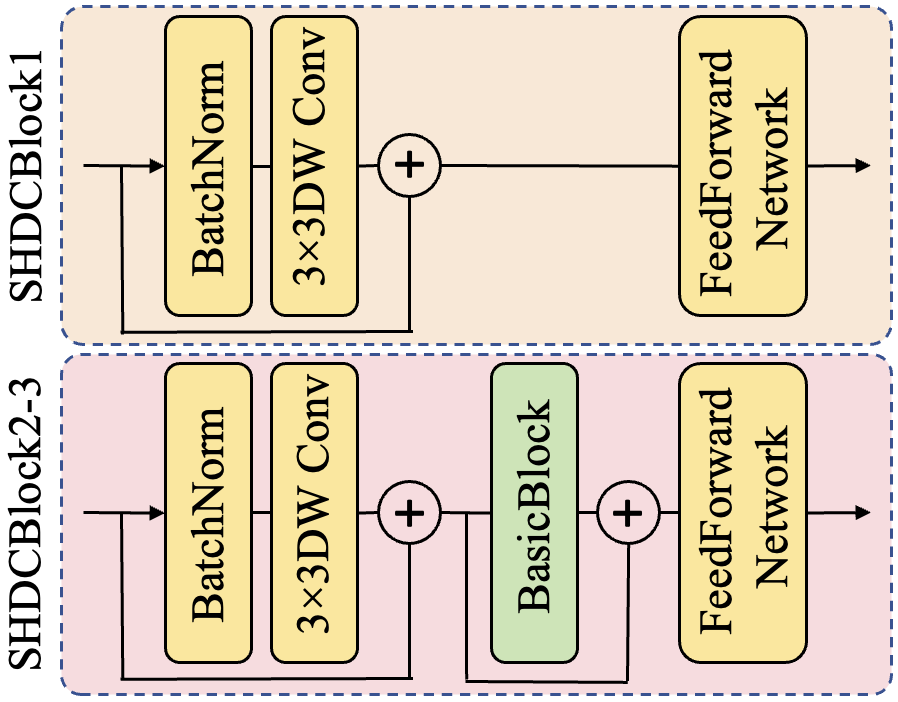}
	\caption{An overview of the encoder architecture. To balance computational efficiency and model performance, the first SHDCBlock only employs depthwise convolutions and FFNs. The following two SHDCBlocks introduce global-local feature fusion modules to enhance multi-scale feature representation.}
	\label{fig:2}
\end{figure}

The encoder is composed of four stages, where the first stage realizes the basic feature extraction via stacking convolutional layers and activation functions. The SHDCBlock is introduced in the later stages, containing depthwise convolutions, the BasicBlock module, and feedforward neural networks (FNNs), to achieve higher-level feature representation. Targeting computational efficiency, as depicted in Fig.~\ref{fig:2}, the first SHDCBlock only uses depthwise convolutions and FNNs. While the subsequent two introduce global and local feature fusion modules to enhance multi-scale feature modeling capability. Along the feature extraction process, the channel dimension increases gradually to capture information at different semantic scales step by step. The decoder process is symmetric to the encoder and includes four upsampling modules. Each layer fuses feature maps from the corresponding scale in the encoder, and realizes progressive multi-scale feature fusion and detailed information recovery by the DyFusionUp module. Finally, the output layer generates a single-channel segmentation map through a 1$\times$1 convolution to achieve the final prediction.

\subsection{Hybrid Feature Extraction Encoder}

\begin{figure}[t]
	\centering
	\includegraphics[width=\linewidth]{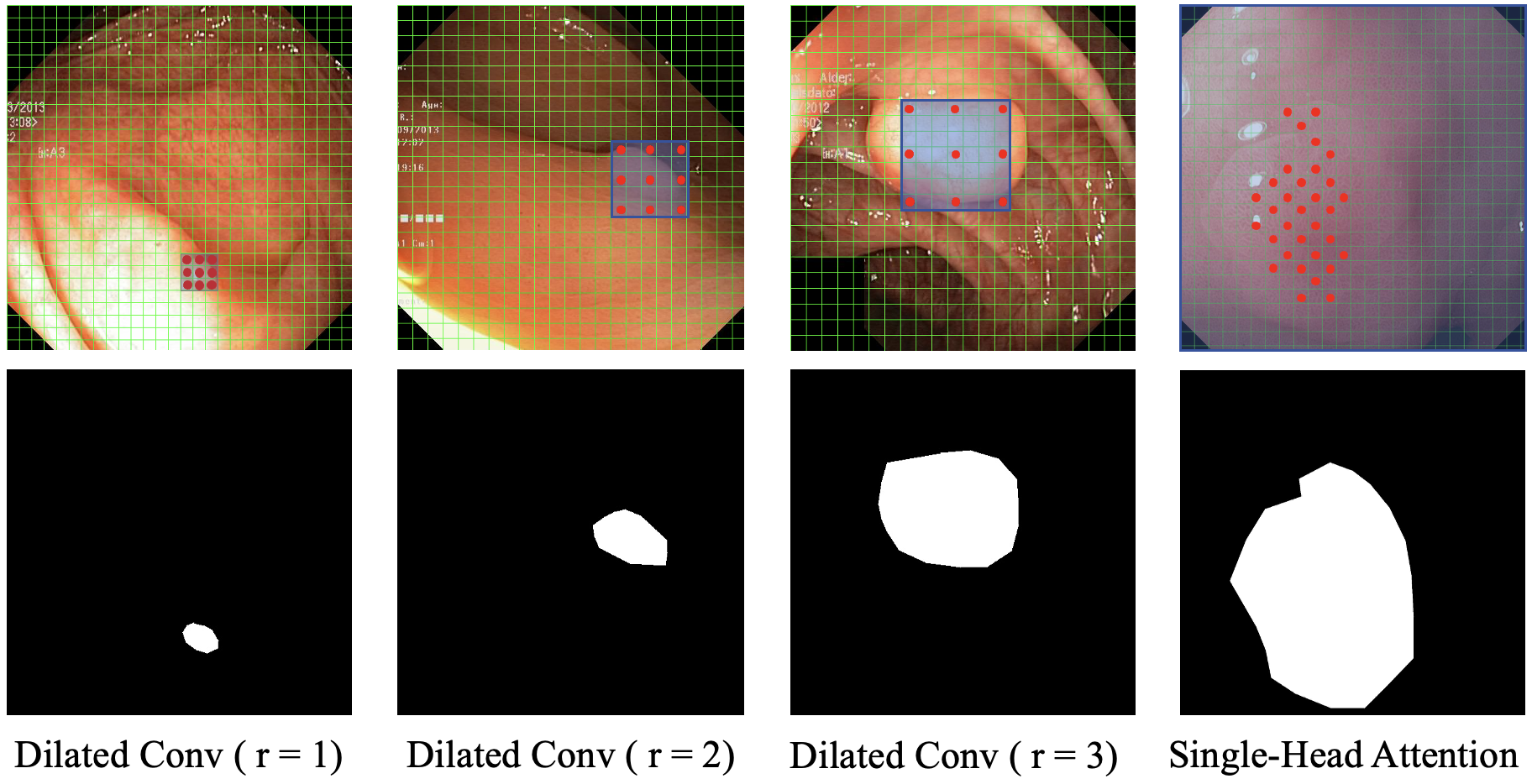}
	\caption{The effect of multi-scale dilated depthwise convolutions. Specifically, we realize various receptive fields over lesion regions via varying dilation rates. As exhibited, dilated convolutions adjust the receptive field size to extract multi-scale contextual features, while the following single-head attention models the long-range dependencies. This provides the modeling across various scales, demonstrating superior semantic perception in irregular lesion boundaries.}
	\label{fig:3}
\end{figure}

Medical image segmentation, especially for small lesions, exhibits multi-scale differences and structural blurriness. Considering gastrointestinal polyps and skin lesions, the former appearances are dominated by large regions with complex shapes, while the latter are generally small regions with blurred boundaries. In such cases, traditional CNNs are limited by their fixed receptive fields and local modeling capabilities; thus, it is difficult for them to capture both global context and fine-grained boundary information simultaneously. To bridge this gap, we propose a hybrid feature extraction encoder, combining the \textbf{S}ingle-\textbf{H}ead self-attention mechanism with multi-scale dilated \textbf{D}epthwise \textbf{C}onvolutions, abbreviated as SHDCBlock, to achieve efficient collaborative modeling of both global and local features, as in Fig.~\ref{fig:3}. Particularly, the DyT (Dynamic Tanh)~\cite{zhu2025transformers} is introduced to replace the layer normalization, adapting to the response of various region features.  

\noindent\textbf{SHDCBlock.} Denoting the input features as $X\in \mathbf{R}^{B\times C\times H\times W}$, we firstly divide the features into two parts along the channel dimension: one part $X_g$ for modeling global relationships and the other part $X_l$ for capturing local context:
\begin{equation}
	X=[X_g, X_l]; \quad X_g\in \mathbf{R}^{B\times C_g\times H\times W}, X_l\in \mathbf{R}^{B\times (C-C_g)\times H\times W}.
\end{equation}

For global modeling, we adopt a lightweight single-head attention mechanism, which realizes cross-spatial context perception by a group of tensors: query (Q), key (K), and value (V):
\begin{gather}
	\phi(X_g)=\gamma\cdot \tanh(\alpha X_g)+\beta,\\
	[Q, K, V]=Conv_{1\times1}(\phi(X_g)),
\end{gather}
where $\phi(x)$ represents the dynamic non-linear transformation realized through the DyT module, and $\alpha, \beta, \gamma$ are all learnable parameters. The attention result is calculated as follows:
\begin{equation}
	\mathrm{Attention}(Q,K,V)=\mathrm{softmax}(\frac{Q^T K}{\sqrt{d}})V.
\end{equation}

Compared with the computationally expensive multi-head attention in the commonly used Transformers, single-head attention can significantly reduce computational costs while maintaining modeling capabilities, thus suitable for global relationship modeling requirements of high-resolution input, especially the medical images. Moreover, in cases of blurred boundaries and small targets, such as the lesions, single-head attention can avoid overfitting to feature patterns of specific scales.

For local context modeling, a multi-scale dilated depthwise convolution structure is employed. Three convolution kernels with different dilation rates are designed to expand the receptive fields and enhance the response to different lesion scales:
\begin{equation}
	L_r=\mathrm{DWConv}^{(r)}_{3\times 3} (X_l),\textbf{ } r\in\{1,2,3\},
\end{equation}
where DWConv denotes the depthwise convolution. The latter multi-scale output fusion process is as follows:
\begin{equation}
	X_l^{\text{out}}=\mathrm{BN}(X_l+\sum_{r=1}^3 L_r),
\end{equation}
where BN denotes the batch-normalization layer. The residual structure preserves the original features, and the different dilation rates enable the model to achieve multi-scale context modeling without significantly increasing the number of parameters, which is particularly important for fine structures in tasks such as skin lesions or nodules. In the fusion stage, we concatenate the outputs of the global path and the local path along the channel dimension, followed by performing feature integration via a lightweight $1\times1$ convolution to achieve non-linear interaction between channels and dimension adjustment:
\begin{equation}
	X_{\mathrm{fused}}=\mathrm{Conv}_{1\times1}([\mathrm{Attention}(Q,K,V); X_l^{\text{out}}]).
\end{equation}

As introduced before, instead of the layer normalization in ViTs, we use the DyT module for dynamic normalization adjustment. DyT avoids relying on preset statistics but automatically adjusts the feature amplitude through learnable parameters, making it more adaptable to scenarios with drastic brightness changes and large tissue differences in medical images (e.g., MRI and CT). In summary, the proposed SHDCBlock, with a global-local dual-path structure, takes into account long-range semantic perception and detailed feature expression. The model combines task characteristics to enhance the capability of modeling multi-scale features and blurred boundaries, and simultaneously adopts a lightweight design to ensure inference efficiency, making it suitable for real-time segmentation applications in resource-constrained environments.

\begin{figure}[t]
	\centering
	\includegraphics[width=\linewidth]{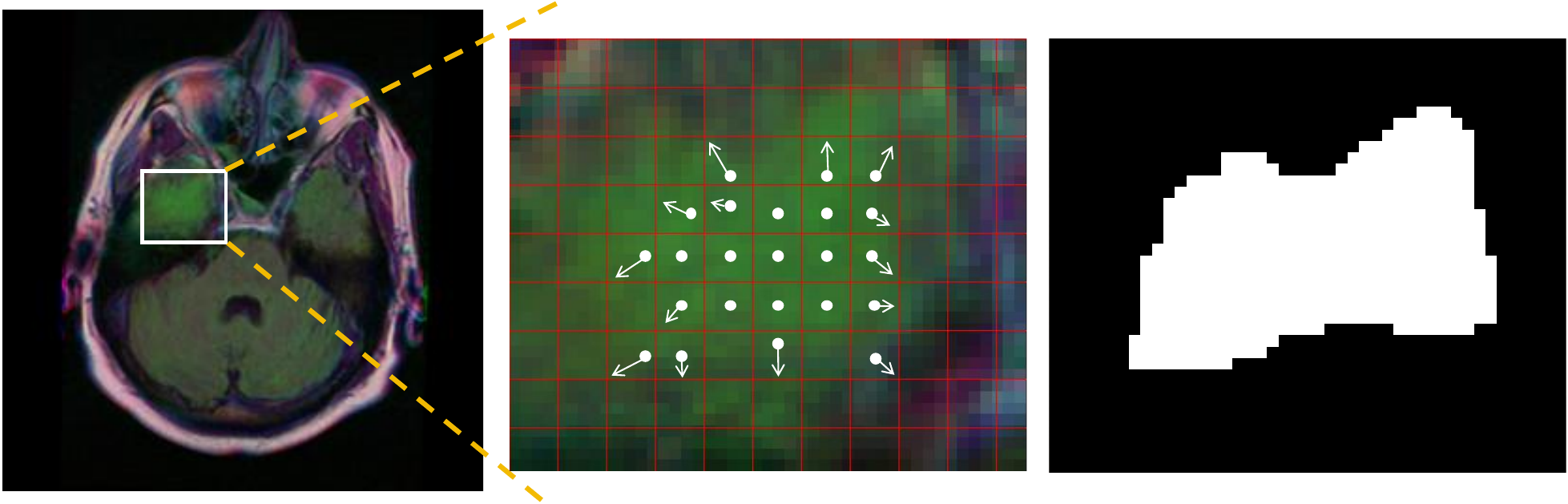}
	\caption{The visualization of dynamically adjusted sampling points with learned offsets within the region of interest (ROI). The directional arrows illustrate both the magnitude and orientation of the adaptive offsets, which are adaptive and helpful to enhance the edge feature extraction.}
	\label{fig:4}
\end{figure}

\subsection{Lightweight Dynamic Adaptive Upsampling Decoder}
In medical image segmentation involving high-resolution images and the identification of tiny lesions, traditional upsampling methods (bilinear interpolation and transposed convolution) fail in edge blurring, detail loss, and checkerboard artifacts, making it hard to guarantee semantic consistency and structural accuracy. To alleviate these issues, we design a DyFusionUp module by introducing a \textbf{Dy}namic sampling mechanism and an efficient \textbf{Fusion} strategy, achieving high-fidelity structure reconstruction and refined edge enhancement.

DyFusionUp is mainly composed of the dynamic upsampling, the channel-alignment convolution, and the fusion-enhancement module, as depicted in Fig.~\ref{fig:pipeline}. To enhance boundary restoration and fine-detail reconstruction in the upsampling process, we employ an adaptive resampling strategy that dynamically refines the sampling grids at a sub-pixel level. Specifically, the upsampling process begins by establishing a set of initial sampling offsets, denoted as $p_{\text{init}}$, which uniformly subdivide each pixel into a 2$\times$2 grid. These offsets are predefined as:
\begin{align}
	p_{\text{init}}=\{(-0.25,-0.25),(-0.25,0.25),\nonumber\\
	(0.25,-0.25),(0.25,0.25)\}
\end{align}
which represent the relative positions of the sub-pixel centers within each original pixel. This initialization ensures that, in the absence of learned displacements, the upsampling operation aligns with standard bilinear interpolation.

Given an input feature map $x\in \mathbf{R}^{C\times H\times W}$, an offset feature map $\Delta p\in \mathbf{R}^{2gH\times W}$ is predicted through a 1$\times$1 convolution layer, where $g$ is the number of channel groups. The predicted offsets are then modulated by a fixed scaling factor to constrain the sampling displacement range, yielding a refined offset $\Delta p'$,
\begin{gather}
	\Delta p'=0.25\times \Delta p, \\
	G(p)=\frac{2(p_{\text{init}}+\Delta p')}{[W, H]}-1,\\
	\hat{x}(p)=\sum_{q\in \mathbf{N}(G(p))}\omega(q) \cdot x(q),
\end{gather}
where $G(p)$ denotes the normalized sampling coordinate, $\mathbf{N}$ stands for the interpolation neighborhood, $\omega(\cdot)$ is the interpolation weight, and $\hat{x}(p)$ represents the output. During the sampling process, features are processed in a group-wise manner to reduce computation. As depicted in Fig.~\ref{fig:4}, the dynamic upsampling provides an adaptive sampling grid, which has better spatial adaptability, thus effectively enhancing boundary recovery and detail restoration.  

Then, the up-sampled features $F_u\in \mathbf{R}^{B\times C'\times H\times W}$ are aligned with channel dimension of the skip-connected features $F_s\in \mathbf{R}^{B\times C\times H\times W}$ through the $1\times 1$ convolution transformation:
\begin{equation}
	F_u^{\text{aligned}}=\mathrm{Conv}_{1\times 1}(F_u).
\end{equation}
Next, the concatenated features $F_{\text{cat}}=[F_s, F_u^{\text{aligned}}]$ are fed into a lightweight multi-scale dilated convolution module to further enable cross-channel enhancement and context aggregation. Moreover, to extract local structures, a 3$\times$3 convolution is attached at the end for spatial information fusion. 


The DyFusionUp module realizes intelligent up-sampling via a dynamic offset mechanism, significantly outperforming traditional methods. The structurally adaptive characteristic enables precise adaptation to changes in edge morphology and lesion contours, particularly adept at handling low-contrast regions (e.g., colonic polyps and skin lesions). Further, by adopting grouped operations and lightweight convolutions, it substantially cuts down the computational bottleneck while maintaining performance, making it well-suited for scenarios with limited resources. With integrated features from multi-scale skip connections, the decoder gradually restores spatial resolution and offers synergistic optimization of semantic information and boundary features for the final segmentation task.

\subsection{Loss Function}

To mitigate class imbalance and boundary ambiguity issues, we employ a widely used effective hybrid loss function~\cite{feng2020cpfnet,dai2024i2u}. Specifically, we adopt a combination of Dice loss and binary cross-entropy (BCE) loss, where the Dice loss emphasizes regional overlap and mitigates foreground-background imbalance. In contrast, the BCE loss enhances pixel-level classification by focusing on boundary precision. Thus, the model effectively balances sensitivity to spatial overlap with fine-grained classification accuracy. The formulation of the Dice loss is as follows:
\begin{equation}
	\mathcal{L}_{\text{Dice}}=1-\frac{2\sum_i p_i\cdot g_i +\epsilon}{\sum_i p_i+\sum_i g_i +\epsilon},
\end{equation}
where $p_i$ and $g_i$ represent the predicted probability and ground truth label of the i-th pixel, respectively, and $\epsilon$ is the introduced smoothing term. The incorporated BCE loss is defined as:
\begin{equation}
	\mathcal{L}_{\text{CE}}=-\frac{1}{N}\sum_i [g_i \log \sigma(p_i)+(1-g_i)\log(1-\sigma(p_i))].
\end{equation}
Then the final loss function is a weighted sum of these two components, controlled by a balancing coefficient $\lambda$:
\begin{equation}
	\mathcal{L}=\lambda \cdot \mathcal{L}_{\text{CE}} + (1-\lambda)\cdot \mathcal{L}_{\text{Dice}}.
\end{equation}
In experiments, we empirically set $\lambda=0.5$~\cite{valanarasu2021medical}, which achieves a satisfactory balance between segmentation accuracy and training stability.

\section{Experimental Results}

\begin{table*}
	\centering
	\caption{The characteristics of the experimental datasets. Note that N/A indicates the split does not contain that part.}
		\begin{tabular}{lllllll}
			\toprule
			\textbf{Dataset} & \textbf{Count} & \textbf{Resize} & \textbf{Train} & \textbf{Valid} & \textbf{Test} & \textbf{Modality} \\
			\midrule
			Kvasir-SEG & 1000 & (224,224) & 800 & 100 & 100 & Colon polyp images \\ 
			CVC-ClinicDB & 612 & (224,224) & 489 & 61 & 62 & Colon polyp images \\ 
			Brain-MRI & 1373 & (224,224) & 1089 & 137 & 138 & MR Images \\ 
			ISIC2016 & 1279 & (224,224) & 900 & N/A & 379 & Dermoscopy Images \\ 
			PH2 & 200 & (224,224) & 160 & 20 & 20 & Dermoscopy Images \\ 
			GlaS & 165 & (224,224) & 132 & 16 & 17 & Colorectal Cancer Slides \\ 
			TNBC & 50 & (224,224) & 40 & N/A & 10 & Breast Cancer Tissue Sections \\ 
			\bottomrule
	\end{tabular}
	\label{tab:dataset}
\end{table*}
    
\subsection{Datasets and Experimental Setup}
\textit{Datasets.} We conduct comprehensive evaluations on the following seven datasets, whose characteristics are summarized in Table~\ref{tab:dataset}. Next, we introduce each dataset in detail.

\noindent\textbf{Kvasir-SEG} dataset~\cite{jha2019kvasir,jha2019resunet++} contains $1000$ professionally annotated images of colon polyps with different image resolutions. In experiments, the dataset is divided with 8:1:1 ratio, where 80$\%$ is used as the training set, and the remaining 20$\%$ is evenly split into the validation and testing sets to ensure the reliability of model evaluation.

\noindent\textbf{CVC-ClinicDB}~\cite{bernal2015wm} dataset is a publicly available dataset containing $612$ frame images extracted from colonoscopy videos. Each image is of $384\times288$ and precise pixel-level annotations of polyp regions are provided.


\noindent\textbf{Brain-MRI}~\cite{buda2019association} dataset, which is dedicated to brain tumor segmentation, contains $3,929$ brain MRI scans from $110$ patients acquired using the FLAIR sequence. For data division, we choose $1,373$ samples with tumor, and randomly divide into the training, validation, and testing sets with 8:1:1 ratio.

\noindent\textbf{ISIC2016}~\cite{gutman2016skin} skin lesion segmentation dataset contains $1,279$ dermoscopic images with pixel-level annotated lesion region masks, and is split with $900$ images for training and $379$ for testing. The dataset is a commonly used benchmark, especially suitable for the research and evaluation of automatic skin lesion segmentation algorithms.

\noindent\textbf{PH2}~\cite{mendoncca2013ph} dataset contains $200$ high-resolution dermoscopic images. All images are annotated by professional physicians with precise lesion region segmentation masks and clinical diagnosis labels.
The whole dataset is divided into training, validation, and testing set with 8:1:1 ratio, providing a reliable benchmark for model evaluation.

\noindent\textbf{GlaS}~\cite{sirinukunwattana2017gland} dataset contains $165$ images derived from H\&E-stained colorectal adenocarcinoma (stage T3 or T42) tissue sections of $16$ patients. The whole dataset is randomly divided into the training, validation, and testing set with 8:1:1 ratio.

\noindent\textbf{TNBC}~\cite{naylor2018segmentation} dataset contains $50$ breast cancer tissue section images and is divided into the training and testing sets with 4:1 ratio. The dataset is used for cell analysis of breast cancer and has important applications, especially in breast cancer molecular pathology and tumor microenvironment. 

\textit{Implementation Details.} 
All models are implemented based on the PyTorch framework, and experiments are conducted on an NVIDIA RTX 4090 GPU card. The input images are uniformly resized to $224\times224$. To enhance the model generalization ability, various data augmentation techniques are applied during training, including random cropping (with the cropped image size of $224\times224$ and a scaling ratio ranging from $0.5$ to $1.0$), horizontal flipping \& vertical flipping (both with the probability of $0.5$), random rotation (with some rotation angle ranging in $[-15^\circ, 15^\circ]$ and the probability of $0.6$), elastic deformation (with the probability of $0.3$), and brightness and contrast perturbation (with the probability of $0.2$). Additionally, all images are normalized with the mean of $[0.485, 0.456, 0.406]$ and the standard deviation of $[0.229, 0.224, 0.225]$.

During training, we use the AdamW optimizer with $\beta$ values set to $(0.9, 0.999)$, and gradient clipping is implemented. The initial learning rate is set as $1e-3$ with the weight decay coefficient of $3e-5$. A polynomial decay strategy is adopted for the learning rate scheduler, and a linear warm-up phase of $10$ epochs is set. The total number of training epochs is $130$. To accelerate the training process, mixed-precision computing is enabled. The batch size is set to $16$, and the random seed is fixed at 42 to ensure the reproducibility of the experiments.

\textit{Evaluation Metrics.} To comprehensively evaluate the model performances of the proposed model, we employ six widely recognized quantitative evaluation metrics (Dice, IoU, Precision, Recall, Specificity, Accuracy) for systematic analysis. These metrics quantify the segmentation results from two dimensions: regional overlap and classification accuracy.

\textit{Comparison Methods.} To ensure comprehensive benchmarking, we compare our model with various state-of-the-art methods, including CNN-based, transformer-based, and hybrid architectures. For fair comparison, all the methods are trained under the same implementation settings (epochs, losses, and optimizer). The CNN-based methods include UNet~\cite{ronneberger2015u}, AttentionUNet~\cite{oktay2018attention}, Unet++~\cite{zhou2019unet++}, DCSAUNet~\cite{xu2023dcsau}, M2SNet~\cite{zhao2023m}, and CMUNeXt-Large~\cite{tang2024cmunext}. Transformer-based methods consist of MISSFormer~\cite{huang2021missformer} and SwinUNet~\cite{cao2022swin}, while the hybrid methods encompass HiFormer-Base~\cite{heidari2023hiformer}, H2Former~\cite{he2023h2former}, FATNet~\cite{wu2022fat}, and BEFUnet~\cite{manzari2024befunet}.

\subsection{Segmentation Results Analysis}

\textit{Polyp Segmentation.} We first evaluate the performance of our proposed DyGLNet on two publicly available polyp segmentation datasets, Kvasir-SEG and CVC-ClinicDB, and conduct a comparative analysis with current mainstream methods. As shown in Table~\ref{tab:2}, on the Kvasir-SEG dataset, our DyGLNet reaches 91.34\% Dice and 85.55\% IoU, outperforming the CNN-based CMUNeXt-L by 2.01\% and 3.18\%. Similarly, our model gains 1.76\% and 2.81\% improvements over the Transformer-based SwinUnet on Dice and IoU. Moreover, DyGLNet surpasses the SOTA method FATNet, which also adopts a hybrid CNN-Transformer network design.
A similar phenomenon can also be observed on the CVC-ClinicDB dataset, and DyGLNet further increases the Dice to 93.71\% and IoU to 88.77\%, outperforming all the other methods, indicating the effectiveness of our model.

\begin{table*}
\centering
\caption{Comparisons on the Kvasir-SEG \& CVC-ClinicDB polyp segmentation datasets. The best and second-best results are marked by \textcolor{red}{red} and \textcolor{blue}{blue}, respectively.}
\begin{tabular}{cclcccccc}
	\toprule
	&&Model & Dice  & IoU   & Prec & Recall & Spec & Acc\\ 
	\midrule
	\multirow{13}{*}{\rotatebox{90}{Kvasir-SEG dataset}} & \multirow{6}{*}{CNN-based} & Unet & 87.96 & 80.92 & 89.52 & 90.17 & 97.90 & 96.64\\ 
	&&AttentionUnet & 88.36 & 81.60 & 89.75 & 90.52 & 97.86 & 96.46\\ 
	&&Unet++ & 88.77 & 81.58 & 88.86 & 91.51 & 97.80 & 96.69\\ 
	&&DCSAUnet & 87.37 & 81.14 & 89.60 & 88.12 & 97.93 & 96.57\\ 
	&&M2SNet & 83.57 & 75.86 & 85.80 & 87.28 & 97.61 & 95.78\\ 
	&&CMUNeXt-L & 89.54 & 82.91 & 90.32 & 91.62 & 97.99 & 96.53\\ 
	\cmidrule{2-9}
	&\multirow{2}{*}{Trans-based} & MISSFormer & 85.37 & 78.05 & 85.04 & 88.66 & 97.19 & 95.86\\ 
	&& SwinUnet & 89.76 & 83.21 & 90.13 & 92.26 & 97.65 & 96.88\\ 
	\cmidrule{2-9}
	&\multirow{5}{*}{Hybrid} & HiFormer-B & 89.50 & 83.46 & 90.55 & 91.38 & 97.73 & 96.91\\ 
	&&H2Former & 90.18 & 84.05 & \textcolor{blue}{90.92} & 92.20 & 97.88 & 96.83\\ 
	&&FATNet & \textcolor{blue}{90.58} & \textcolor{blue}{84.30} & 90.34 & \textcolor{blue}{92.86} & \textcolor{blue}{98.15} & \textcolor{blue}{97.21}\\ 
	&&BEFUnet & 85.60 & 77.79 & 86.03 & 89.78 & 97.14 & 96.10\\ 
	\rowcolor{gray!20}&&DyGLNet & \textcolor{red}{91.34} & \textcolor{red}{85.55} & \textcolor{red}{91.59} & \textcolor{red}{92.97} & \textcolor{red}{98.33} & \textcolor{red}{97.34}\\ 
	\midrule
	\midrule
	\multirow{13}{*}{\rotatebox{90}{CVC-ClinicDB dataset}}&\multirow{6}{*}{CNN-based}&Unet  & 90.28 & 84.08 & 92.01 & 90.61 & 99.41 & 98.75\\ 
	&&AttentionUnet & 91.48 & 85.44 & 91.70 & 92.77 & 99.35 & 98.90\\ 
	&&Unet++ & 90.77 & 85.01 & 91.84 & 92.41 & 99.28 & 98.84\\ 
	&&DCSAUnet & 88.35 & 82.51 & 89.00 & 89.61 & 99.26 & 98.81\\ 
	&&M2SNet & 89.40 & 83.62 & 88.54 & 90.92 & 99.20 & 98.76\\ 
	&&CMUNeXt-L & 90.96 & 84.97 & 90.22 & 92.74 & 99.13 & 98.80\\ 
	\cmidrule{2-9}
	&\multirow{2}{*}{Trans-based} & MISSFormer & 90.01 & 82.97 & 91.78 & 90.16 & 99.33 & 98.83\\ 
	&&SwinUnet & \textcolor{blue}{92.70} & 86.94 & 92.53 & 93.61 & 99.43 & 99.06\\ 
	\cmidrule{2-9}
	&\multirow{5}{*}{Hybrid} & HiFormer-B & 92.45 & 87.48 & \textcolor{red}{93.98} & 92.58 & 99.45 & 99.06\\
	&&H2Former & 92.49 & \textcolor{blue}{87.86} & 92.70 & 92.72 & \textcolor{blue}{99.50} & \textcolor{blue}{99.09}\\
	&&FATNet & 91.60 & 85.43 & 90.47 & \textcolor{blue}{93.90} & 99.23 & 98.93\\
	&&BEFUnet & 88.74 & 81.95 & 89.19 & 89.88 & 99.32 & 98.82\\
	\rowcolor{gray!20}& & DyGLNet & \textcolor{red}{93.71} & \textcolor{red}{88.77} & \textcolor{blue}{93.50} & \textcolor{red}{94.59} & \textcolor{red}{99.52} & \textcolor{red}{99.24}\\
	\bottomrule
\end{tabular}%
\label{tab:2}
\end{table*}%

\begin{figure*}
\centering
\includegraphics[width=0.99\linewidth]{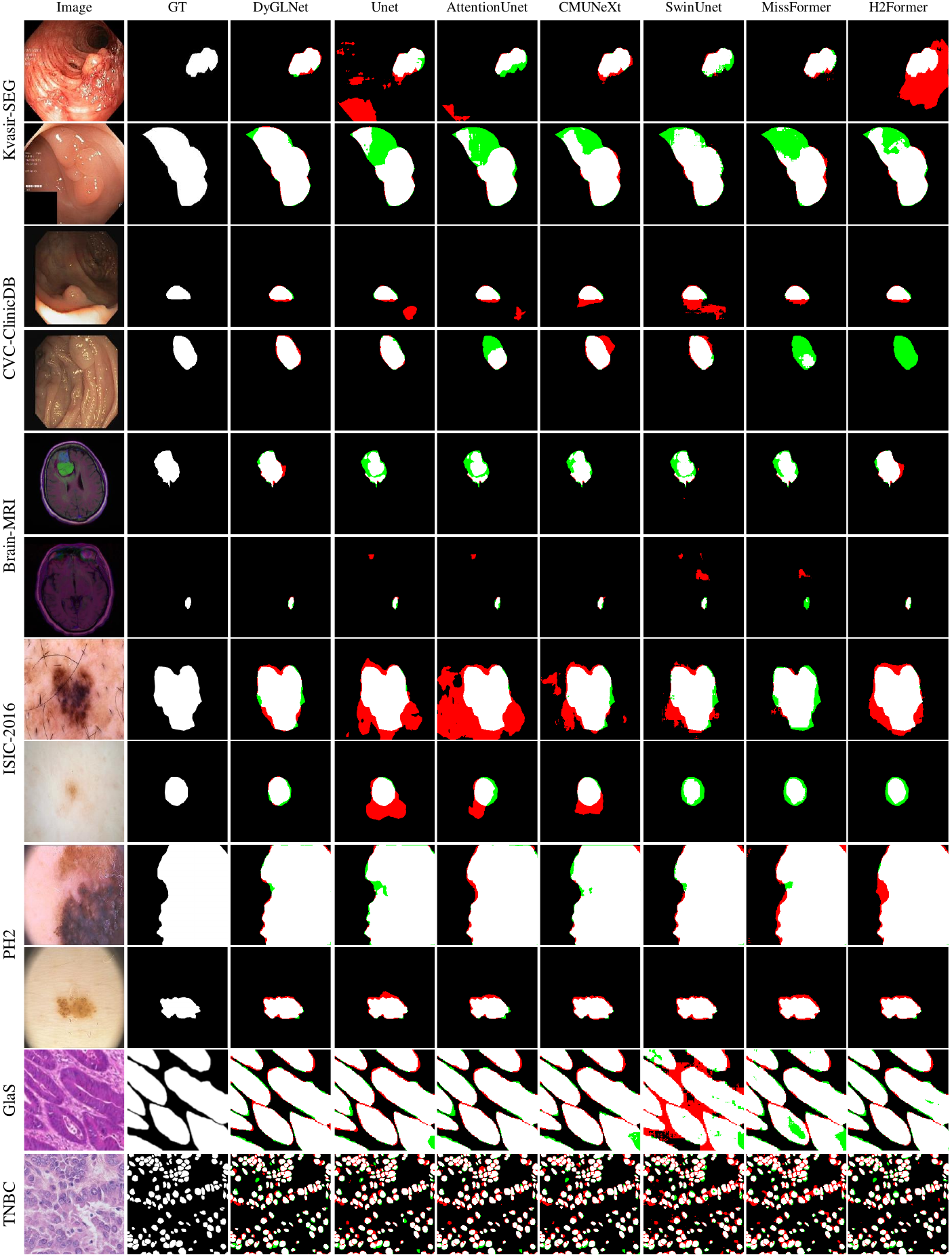}
\caption{Visualization of Medical image segmentation results. Red regions indicate over-segmented areas, green regions denote under-segmented areas.}
\label{fig:visual}
\end{figure*}

To further compare the segmentation results between DyGLNet and other methods, we visualize predicted masks on both datasets in Fig.~\ref{fig:visual}. For polyp images with blurred boundaries (such as cases in Kvasir-SEG), the segmentation contours predicted by DyGLNet fit the annotations better, while the CNN-based UNet and Transformer-based SwinUnet have problems, such as blurred edges or local breaks. Although H2Former exhausts to combine CNNs and ViTs, the predictions fail to provide accurate contours. Further, for tiny polyps (with a diameter$<$5mm), with the help of dynamic up-sampling, DyGLNet completely preserves the lesion shapes, while other methods suffer from missed detection or discontinuous regions. Therefore, with a considerable combination of the Transformers and dilated convolution blocks, DyGLNet can encode both local features and long-range dependencies, enabling predictions to be more precise for both large and tiny polyps.

\textit{Brain Tumor Segmentation.}
Then we test our model on the Brain-MRI dataset, and compare with SOTA methods in Table~\ref{tab:3}. DyGLNet achieves 88.52\% Dice coefficient and 81.12\% IoU, which consistently outperforms other comparative methods. Particularly, hybrid models not only surpass CNN-based models but also outperform Transformer-based methods, indicating the advantages of modeling both local and global features. Furthermore, DyGLNet consistently performs better than all other hybrid models with our considerable design of the dynamic feature fusion strategy.
Meanwhile, DyGLNet achieves the highest precision (88.63\%) and specificity (99.76\%), indicating the effectiveness in reducing false-positive predictions and accurately identifying tumor boundaries.
To further validate the model sensitivity, we also compute the recall (89.76\%) and accuracy (99.56\%), and DyGLNet outperforms all the models except AttentionUnet and DCSAUnet on the Recall metric. We analyze that this is caused by the conflict between global modeling and the sensitivity, and similar issues can be observed in all the Transformer-based and hybrid models. Fortunately, our employment of dilated convolutions enables DyGLNet to perform almost the best in this category. 

\begin{table*}
\centering
\caption{Comparisons with SOTA models on the Brain-MRI dataset.}
\begin{tabular}{clcccccc}
	\toprule
	& Model & Dice  & IOU  & Prec & Recall & Spec & Acc \\
	\midrule
	\multirow{6}{*}{CNN-based} & Unet  & 86.40 & 78.84 & 86.15 & 88.42 & 99.72 & 99.52 \\
	& AttentionUnet & 86.99 & 79.51 & 85.83 & \textcolor{red}{90.37} & 99.69 & 99.52 \\
	& Unet++ & 86.82 & 79.64 & 85.85 & 88.99 & \textcolor{blue}{99.74} & 99.54\\
	& DCSAUnet & 87.53 & 79.97 & 87.29 & \textcolor{blue}{89.87} & 99.72 & 99.54\\
	& M2SNet & 84.69 & 76.17 & 83.62 & 89.72 & 99.54 & 99.38\\
	& CMUNeXt-L & 86.66 & 79.04 & 86.37 & 89.06 & 99.72 & 99.53 \\
	\midrule
	\multirow{2}{*}{Trans-based} & MISSFormer & 86.57 & 78.53 & 86.82 & 88.11 & 99.73 & 99.49\\
	& SwinUnet & 83.39 & 74.88 & 83.54 & 86.43 & 99.64 & 99.38\\
	\midrule
	\multirow{5}{*}{Hybrid} & HiFormer-B & 87.39 & 79.56 & 86.75 & 89.79 & 99.74 & 99.54\\
	& H2Former & \textcolor{blue}{87.89} & \textcolor{blue}{80.40} & \textcolor{blue}{87.86} & 89.56 & \textcolor{red}{99.76} & \textcolor{red}{99.56}\\
	& FATNet & 86.60 & 79.50 & 87.27 & 88.28 & 99.75 & \textcolor{blue}{99.55}\\
	& BEFUnet & 85.94 & 78.00 & 85.04 & 89.09 & 99.70 & 99.48\\
	\rowcolor{gray!20}& DyGLNet & \textcolor{red}{88.52} & \textcolor{red}{81.12} & \textcolor{red}{88.63} & 89.76 & \textcolor{red}{99.76} & \textcolor{red}{99.56}\\
	\bottomrule
\end{tabular}
\label{tab:3}%
\end{table*}%

What's more, we also visually exhibit comparisons on the dataset in the 5th and 6th rows of Fig.~\ref{fig:visual}. Consistently, our predicted masks better fit the annotated tumors. The better predictions of DyGLNet mainly derive from the SHDCBlock module since it is capable of achieving effective fusion of global context modeling and local feature extraction through the collaborative work of a single-head attention mechanism and multi-scale dilated convolutions. The characteristic is particularly important for brain tumor segmentation as brain lesions have irregular shapes and significant scale differences. The dynamic up-sampling module (DyFusionUp) further enhances the ability to restore edge details, effectively solving the common boundary blur problem in MRI images. In contrast, pure Transformer-based models (such as SwinUnet) perform poorly due to the insufficient local modeling capability, while CNN-based models struggle to capture long-range dependencies in large tumor regions. These results fully demonstrate the necessity of combining global and local features in brain tumor segmentation.

\textit{Dermoscopic Image Segmentation.}
To validate the segmentation robustness on different image modalities, we next conduct experiments on two dermoscopic image datasets, ISIC2016 and PH2, and DyGLNet also exhibits excellent performance in skin lesion segmentation. 
As shown in Table~\ref{tab:4}, on the ISIC2016 dataset, DyGLNet achieves 91.71\% Dice coefficient and 85.78\% IoU. This is comparable to the SOTA BEFUnet (Dice 91.58\%, IoU 85.52\%) and significantly outperforms both CNN-based methods and pure Transformer-based methods. Similarly, on the PH2 dataset, DyGLNet achieves the best precision (96.73\%) and IoU (91.32\%), while its Dice coefficient (95.40\%) and accuracy (98.13\%) also almost outperform other methods, verifying the model's superior performance in distinguishing benign and malignant skin lesions. 
\begin{table*}
\centering
\caption{Comparisons with SOTA models on the ISIC2016 and PH2 Dataset.}
\begin{tabular}{clcccccccc}
	\toprule
	& \multirow{2}[4]{*}{Model} & \multicolumn{4}{c}{ISIC2016}  & \multicolumn{4}{c}{PH2}\\
	\cmidrule{3-10} & & Dice  & IOU   & Prec & Acc & Dice  & IOU   & Prec & Acc \\
	\midrule
	\multirow{6}{*}{CNN-based} & Unet  & 90.75 & 84.36 & 90.51 & 95.60 & 94.35 & 89.46 & 95.35 & 97.71\\
	& AttentionUnet & 90.57 & 84.02 & 91.08 & 95.46 & 94.42 & 89.61 & 96.08 & 97.73\\
	& Unet++ & 90.90 & 84.66 & 91.44 & 95.72 & 93.50 & 88.11 & 92.97 & 96.21\\
	& DCSAUnet & 90.95 & 84.57 & 90.02 & 95.70 & \textcolor{blue}{95.34} & \textcolor{blue}{91.15} & 95.85 & 98.01\\
	& M2SNet & 91.19 & 84.99 & 92.19 & 95.56 & 94.96 & 90.53 & 94.97 & 97.04\\
	& CMUNeXt-L & 90.90 & 84.73 & 91.49 & 95.71 & 95.22 & 91.03 & \textcolor{blue}{96.27} & 98.12\\
	\midrule
	\multirow{2}{*}{Trans-based} & MISSFormer & 90.57 & 84.32 & 91.89 & 95.40 & 90.50 & 85.55 & 88.65 & 94.71\\
	& SwinUnet & 89.94 & 83.18 & 91.46 & 95.18 & 95.23 & 91.04 & 96.11 & \textcolor{red}{98.18}\\
	\midrule
	\multirow{5}{*}{Hybrid}  & HiFormer-B & 91.00 & 84.98 & 92.01 & 95.85 & 95.20 & 90.97 & 94.34 & 98.09\\
	& H2Former & 91.23 & 85.01 & \textcolor{red}{92.64} & 95.88 & 95.21 & 91.00 & 94.68 & 97.30\\
	& FATNet & 90.42 & 83.91 & 90.70 & 95.17 & 95.29 & 91.13 & 95.35 & 97.39\\
	& BEFUnet & \textcolor{blue}{91.58} & \textcolor{blue}{85.52} & \textcolor{blue}{92.45} & \textcolor{blue}{96.02} & 93.80 & 88.71 & 92.07 & 97.81\\
	\rowcolor{gray!20}& DyGLNet & \textcolor{red}{91.71} & \textcolor{red}{85.78} & 91.51 & \textcolor{red}{96.11} & \textcolor{red}{95.40} & \textcolor{red}{91.32} & \textcolor{red}{96.73} & \textcolor{blue}{98.13}\\
	\bottomrule
\end{tabular}%
\label{tab:4}
\end{table*}%

The visual comparisons are also shown in 7th-10th rows of Fig.~\ref{fig:visual}, and the visualizations further confirm the superiority of DyGLNet: in typical cases, the model can accurately segment irregularly shaped melanoma regions (7th, 9th rows) while avoiding misclassifying surrounding normal pigmentation as lesions; for small-sized early lesions (8th, 10th rows), DyGLNet provides better complete segmentations than other methods. These observations fully demonstrate the potential practical application of the proposed model in early skin cancer diagnosis.

\textit{Pathological Tissue Segmentation.}
We also conduct experiments on two pathological tissue segmentation datasets, GlaS and TNBC. As shown in Table~\ref{tab:5}, DyGLNet achieves 94.32\% Dice coefficient and 89.36\% IoU on the GlaS dataset, consistently outperforming other comparative models. Compared with both the CNN-based models and the Transformer-based models, DyGLNet improves segmentation performance while simultaneously maintaining high precision. In addition, DyGLNet achieves the best precision (94.05\%) and accuracy (94.40\%), indicating that the model can accurately identify pathological tissue regions and reduce missegmentation.
On the TNBC dataset, 80.05\% Dice coefficient and 66.93\% IoU are attained by DyGLNet, also outperforming other methods. It is worth noting that although the TNBC dataset has a small sample size (50 images), DyGLNet still demonstrates strong generalization ability, with its precision (82.04\%) and accuracy (95.47\%) ranking first. This result indicates that DyGLNet has significant advantages in the segmentation of complex pathological tissues (such as the breast cancer microenvironment).

\begin{table*}
\centering
\caption{Performance comparisons of pathological tissue segmentation on the GlaS and TNBC Datasets.}
\begin{tabular}{clcccccccc}
	\toprule
	& \multirow{2}[4]{*}{Model} & \multicolumn{4}{c}{Glas}      & \multicolumn{4}{c}{TNBC}\\
	\cmidrule{3-10}    & & Dice  & IOU   & Prec & Acc & Dice  & IOU   & Prec & Acc\\
	\midrule
	\multirow{6}{*}{CNN-based} & Unet  & 92.09 & 85.67 & 90.01 & 92.37 & 78.36 & 64.73 & 80.41 & 95.05\\
	& AttentionUnet & 93.86 & 88.61 & \textcolor{blue}{93.81} & 94.09 & 79.19 & 65.85 & 79.94 & 95.30\\
	& Unet++ & 92.54 & 86.40 & 91.86 & 92.96 & 79.61 & 66.22 & 76.75 & 95.08\\
	& DCSAUnet & 93.01 & 87.13 & 92.59 & 93.14 & 75.92 & 61.40 & 75.24 & 94.33 \\
	& M2SNet & 93.69 & 88.31 & 93.49 & 93.84 & 75.18 & 60.48 & 72.22 & 93.47\\
	& CMUNeXt-L & 92.69 & 86.67 & 93.12 & 93.27 & 77.45 & 63.65 & 79.79 & 95.17\\
	\midrule
	\multirow{2}{*}{Trans-based} & MISSFormer & 91.77 & 84.99 & 91.41 & 91.57 & 72.75 & 57.51 & 71.43 & 93.38\\
	& SwinUnet & 79.75 & 67.06 & 78.78 & 80.00 & 72.87 & 57.75 & 72.61 & 93.29\\
	\midrule
	\multirow{5}{*}{Hybrid} & HiFormer-B & \textcolor{blue}{93.90} & \textcolor{blue}{88.66} & 92.86 & \textcolor{blue}{94.10} & 76.31 & 62.04 & 75.92 & 94.45\\
	& H2Former & 93.33 & 88.07 & 92.41 & 93.90 & \textcolor{blue}{79.79} & \textcolor{blue}{66.50} & \textcolor{blue}{81.47} & \textcolor{blue}{95.42}\\
	& FATNet & 93.08 & 87.26 & 92.84 & 93.16 & 77.12 & 62.94 & 80.19 & 94.95\\
	& BEFUnet & 90.07 & 82.18 & 87.69 & 89.73 & 73.58 & 58.52 & 75.75 & 94.30\\
	\rowcolor{gray!20}& DyGLNet & \textcolor{red}{94.32} & \textcolor{red}{89.36} & \textcolor{red}{94.05} & \textcolor{red}{94.40} & \textcolor{red}{80.05} & \textcolor{red}{66.93} & \textcolor{red}{82.04} & \textcolor{red}{95.47}\\
	\bottomrule
\end{tabular}
\label{tab:5}%
\end{table*}%

As shown in the last two rows of Fig.~\ref{fig:visual}, we compare pathological tissue segmentation masks between DyGLNet and other models. In cases of the GlaS dataset, DyGLNet can accurately segment complex structures of adenocarcinoma tissues (such as glandular morphology), while CNN-based methods (such as UNet) suffer from severe over-segmentation or under-segmentation issues. For the breast cancer cell regions of the TNBC dataset, the segmentation masks of DyGLNet are more consistent with the annotations, especially performing better than other models in cell boundaries and micro-lesion areas. These results further verify the advantages of DyGLNet in pathological image analysis, providing a reliable tool for cancer diagnosis and research.

\begin{table*}[t]
\centering
\caption{Ablation study of the proposed DyGLNet. To analyze its components, we ablate with several variants. The base model DyGLNet achieves the best performance across all datasets, verifying the necessity of each module of the proposed model.}
    \begin{tabular}{clccccccc}
    \toprule
    \multirow{8}{*}{\rotatebox{90}{Dice}}&Model & Kvasir & CVC & Brain-MRI & ISIC2016 & PH2 & GlaS & TNBC \\
    \midrule
    &ours w BiIn & 87.60 & \textcolor{blue}{93.57} & 87.60 &  91.30 & 95.07 & 93.60 & 78.67 \\
    &ours w TConv & 88.66 & 92.61 & \textcolor{blue}{87.89}  & 91.45  & 94.88 & 93.60  & \textcolor{blue}{78.89} \\
    &ours w/o DyT & 88.00 & 92.78 &  87.28  & 91.18  & 95.00  & \textcolor{blue}{94.02}  & 77.88 \\
    &ours w/o DC  & 86.45 & 92.51 & 86.58 & \textcolor{blue}{91.62} & 95.13 & 93.98 & 77.44 \\
    &ours w/o SA  & \textcolor{blue}{89.49} & 91.00 & 87.47 & 91.30 & \textcolor{blue}{95.28} & 89.67 & 78.49\\
    &ours Split-DC & 87.21  & 92.61 & 87.60  & 91.02 & 95.13 & 92.98 & 75.63 \\
    &ours Split-SA & 86.67 & 92.62 & 86.61 & 91.35 & 94.94 & 93.39 & 78.79\\
    \rowcolor{gray!20}&DyGLNet  & \textcolor{red}{91.34} & \textcolor{red}{93.71} & \textcolor{red}{88.52} & \textcolor{red}{91.71} & \textcolor{red}{95.40} & \textcolor{red}{94.32} & \textcolor{red}{80.05}\\
    \midrule
    \multirow{8}{*}{\rotatebox{90}{IOU}}&Model & Kvasir & CVC & Brain-MRI & ISIC2016 & PH2 & GlaS & TNBC \\
    \midrule
    &ours w BiIn &  81.53 & \textcolor{blue}{88.40} & 80.04 & 85.30 & 90.72 & 88.09 & 65.04 \\
    &ours w TConv & 82.30 & 87.14  & \textcolor{blue}{80.44} & 85.43 & 90.38 & 88.09 & \textcolor{blue}{65.37} \\
    &ours w/o DyT & 81.87 & 87.54 & 79.90 & 85.22 & 90.59 & \textcolor{blue}{88.85} & 64.16 \\
    &ours w/o DC  & 79.24 & 87.23 & 78.93 & \textcolor{blue}{85.64} & 90.81 & 88.75 & 63.47 \\
    &ours w/o SA  & \textcolor{blue}{82.98} & 85.49 & 80.07 & 85.39 & \textcolor{blue}{91.12} & 81.79  & 64.80 \\
    &ours Split-DC & 80.12 & 86.90 & 79.88 & 84.73 & 90.86 & 87.08 & 61.37 \\
    &ours Split-SA & 79.51 & 86.84 & 78.76 & 85.23 & 90.50 & 87.75 & 65.17 \\
    \rowcolor{gray!20}&DyGLNet  &  \textcolor{red}{85.55} & \textcolor{red}{88.77} & \textcolor{red}{81.12} & \textcolor{red}{85.78} & \textcolor{red}{91.32} & \textcolor{red}{89.36} & \textcolor{red}{66.93} \\
    \bottomrule
    \end{tabular}%
\label{tab:6}%
\end{table*}%

\subsection{Ablation Study}

To verify the specific contributions of each key module in the proposed model, this study conducts comprehensive ablation experiments on all involved datasets, respectively comparing different settings of normalization, dynamic upsampling, multi-scale dilated convolutions, single-head self-attention, and channel division strategies. The detailed results are reported in Table~\ref{tab:6}, and the final DyGLNet achieves the best performance across all datasets, verifying the necessity of each module of the proposed model. 
More specifically, we first replace the DyT module with ordinary normalization (w/o DyT), and the model performance decreases on all datasets. Especially on the Kvasir-SEG and PH2 datasets, the Dice coefficient drops by 3.34\% and 0.40\%, respectively, indicating that the dynamically adjusting feature responses offered by the DyT module enhance the model's discriminative ability. Further, when replacing dynamic upsampling with bilinear interpolation (w BiIn) and transposed convolution (w TConv), model performances both degrade, suggesting that the adaptivity of the dynamic upsampling mechanism better preserves high-resolution details and structural features.

In the downsampling stage, removing three dilated convolutions (w/o DC) also results in performance degradation, especially on the Brain-MRI dataset, indicating that multi-scale receptive fields play a crucial role in localizing complex morphological target regions. Similarly, removing the single-head self-attention module (w/o SA) also causes performance degradation, demonstrating that the module contributes to the long-range dependencies within local regions. When ablating on channel division strategies, allocating more channels to both the dilated convolution branch (Split-DC) or the single-head attention branch (Split-SA) leads to worse performances on the Kvasir-SEG and Brain-MRI datasets, indicating that a reasonable channel division structure facilitates collaborative modeling between different feature branches. In contrast, simply strengthening either branch is prone to causing redundancy or feature information loss.
Based on the above observations, all proposed modules are indispensable and work together to enhance the feature expression ability and segmentation performances, leading to optimal results of the final DyGLNet.

\subsection{Performance versus Computation Analysis}
To comprehensively evaluate the performance of the DyGLNet, we calculate the average Dice (mDice) for models across seven public datasets and show the results in Fig.~\ref{fig:1}. 
The lightest yellow bubble indicates that DyGLNet outperforms other comparative models, demonstrating its stability and generalization ability in various scenarios. 
On the other hand, we compare the computational costs (number of parameters: \#Param, and Flops) of all the models, as shown in Table~\ref{tab:7}.
DyGLNet exhibits significantly better computational efficiency (fewer parameters and lower FLOPs) than other high-performance models, as shown in the above tables. For example, H2Former also performs better than other models, while it requires 24.67G FLOPs, slowing down the inference. In contrast, DyGLNet only requires 11.16G, indicating that it is more suitable for real-world scenarios with limited resources. This advantage benefits from the lightweight single-head attention mechanisms and dynamic upsampling operations.

\begin{table*}[t]
\centering
\caption{Comparisons of the computational costs. The best and second-best results are marked by \textcolor{red}{red} and \textcolor{blue}{blue}, respectively.}
\begin{tabular}{clccc}
	\toprule
         & Model & mDice & \#Param & Flops\\
	\midrule
	\multirow{6}{*}{CNN-based} & Unet & 88.60 & 31.04M & 41.48G \\
	& AttentionUnet & 89.27 & 34.88M & 51.02G \\
	&Unet++ & 88.99 & 9.16M & 26.72G \\
	&DCSAUnet & 88.35 & \textcolor{red}{2.60M} & \textcolor{red}{5.29G}\\
	&M2SNet & 87.53 & 27.69M & 6.91G \\
	&CMUNeXt-L & 89.06 & \textcolor{blue}{8.29M} & 13.16G \\
	\midrule
	\multirow{2}{*}{Trans-based} & MISSFormer & 86.79 & 35.45M & 7.25G \\
        & SwinUnet & 86.23 & 27.15M & \textcolor{blue}{5.91G}\\  
	\midrule
	\multirow{5}{*}{Hybrid} & HiFormer-B & 89.39 & 34.14M & 17.75G \\
	& H2Former & \textcolor{blue}{90.02} & 33.63M & 24.67G \\
	& FATNet & 89.24 & 29.62M & 42.80G\\
	& BEFUnet & 87.04 & 42.61M & 7.95G \\
	\rowcolor{gray!20}& DyGLNet & \textcolor{red}{90.72} & 9.98M & 11.16G \\
	\bottomrule
\end{tabular}%
\label{tab:7}
\end{table*}%

\subsection{Discussions and Limitations}
The DyGLNet proposed in this study demonstrates excellent performance in multiple challenging medical image segmentation tasks through global-local feature fusion and dynamic upsampling mechanisms. Next, we discuss three core aspects. 
1) the SHDCBlock module achieves efficient modeling of multi-scale lesions through the collaboration of single-head self-attention and multi-scale dilated convolutions. Experimental results (on the Kvasir-SEG and CVC-ClinicDB datasets, DyGLNet achieves 91.34\% and 93.71\% Dice coefficients, respectively) verify that the module retains both the local feature extraction capability of CNNs and the model capability to model long-range dependencies, particularly dealing better with blurred boundaries or small targets.
2) the DyFusionUp module realizes adaptive reconstruction of feature maps through a dynamic offset mechanism, significantly improving boundary accuracy segmentation. Compared with bilinear interpolation and transposed convolution, DyFusionUp can dynamically adjust sampling positions according to image content, thus more accurately restoring lesion contours. This characteristic is particularly critical in Brain-MRI and dermoscopic image segmentation tasks since they require lower false positive predictions (i.e., higher precision and specificity) to avoid misdiagnosis.
3). DyGLNet achieves the balance between computational efficiency and model performance. Compared with existing high-performance methods, DyGLNet significantly reduces the number of parameters and computations (only 9.98M parameters and 11.16G FLOPs) while maintaining excellent performance. This makes it more suitable for deployment in medical devices with limited resources.

However, there are still certain limitations. Especially for low-contrast images (such as some MRI sequences), the model still fails to accurately segment lesions. On the other hand, the computation efficiency of the dynamic upsampling module requires consideration when facing real-time scenarios. We would explore more lightweight attention mechanisms or introduce prior knowledge to further enhance model robustness.

\section{Conclusion}
This paper proposes the novel DyGLNet and significantly improves the segmentation accuracy and efficiency by fusing global-local features and dynamic upsampling mechanisms. The core innovations lie in: (1) the SHDCBlock module, which combines single-head self-attention and multi-scale dilated convolutions to achieve long-range dependency and local detail modeling; (2) the DyFusionUp module, which adaptively reconstructs feature maps through a dynamic offset mechanism and effectively addresses the boundary blur problem caused by traditional upsampling methods; (3) a lightweight design that significantly reduces computation complexity while ensuring performances. Comprehensive experiments on several public datasets verify that model performances in various segmentation tasks and the necessity of these designs.
Thus, DyGLNet provides an accurate and efficient solution for medical image segmentation and reveals the potential for clinical applications. In future work, we will further optimize model efficiency and explore its generalization ability for more image modalities (such as ultrasound or pathological sections).

%
%
%
%

\section*{Acknowledgments}

This work is partially supported by the Natural Science Foundation of Henan Province under Grant 222300420140, the National Natural Science Foundation of China under Grants No. 62301532 and 62303438, and in part by the Natural Science Foundation of Jiangsu Province under Grant No. BK20230282.

%

%
%
%

\bibliographystyle{elsarticle-num} 
\bibliography{bio}



%
%
%
\end{document}